\documentclass[review]{elsarticle}
\usepackage{lineno,hyperref}
\modulolinenumbers[5]
\journal{Journal of \LaTeX\ Templates}
\graphicspath{{./images_color/}}
\usepackage{algorithm}
\usepackage[noend]{algpseudocode}
\usepackage{amssymb}
\usepackage{svg}
\usepackage{pifont}
\usepackage{textcomp}
\usepackage{gensymb}
\biboptions{sort&compress}
\newcommand{\cmark}{\ding{51}}%
\newcommand{\xmark}{\ding{55}}%
\begin{document}
\begin{frontmatter}
\title{Operational thermal load forecasting in district heating networks using machine learning and expert advice}


\author[vito,energyville]{Davy Geysen\corref{correspondingauthor}}
\cortext[correspondingauthor]{Corresponding author}
\ead{davy.geysen@vito.be}
\author[vito,energyville]{Oscar De Somer}
\author[noda]{Christian Johansson}
\author[noda]{Jens Brage}
\author[vito,energyville]{Dirk Vanhoudt}
\address[vito]{VITO, Boeretang 200, 2400 Mol, Belgium}
\address[energyville]{EnergyVille, Thor Park 8310, 3600 Genk, Belgium}
\address[noda]{NODA, Biblioteksgatan 4, 374 35 Karlshamn, Sweden}
\begin{abstract}
Forecasting thermal load is a key component for the majority of optimization solutions for controlling district heating and cooling systems. Recent studies have analysed the results of a number of data-driven methods applied to thermal load forecasting, this paper presents the results of combining a collection of these individual methods in an expert system. The expert system will combine multiple thermal load forecasts in a way that it always tracks the best expert in the system. This solution is tested and validated using a thermal load dataset of 27 months obtained from 10 residential buildings located in Rottne, Sweden together with outdoor temperature information received from a weather forecast service. The expert system is composed of the following data-driven methods: linear regression, extremely randomized trees regression, feed-forward neural network and support vector machine. The results of the proposed solution are compared with the results of the individual methods.
\end{abstract}


\end{frontmatter}


\section{Introduction}
A key component of enhancing the energy efficiency in current 3\textsuperscript{rd} and innovative 4\textsuperscript{th} generation district heating systems (DHS) is the ability of predicting the network's future behaviour. More in particular forecasting the thermal load in the system in order to further optimise the DHS controller. Therefore, the work in this paper is an essential part of the Horizon 2020 STORM project in which a generic district heating and cooling network controller needs to be developed \cite{vanhoudt}. 

In recent years, a number of different thermal load forecasting approaches have been investigated, ranging from models solely using historic load data up to more complicated ones incorporating additional parameters like occupancy, meteorological data or physical details of the building. These approaches can be divided in two major classes: the forward and data-driven methods \cite{zhao,fumo}. In the forward or expert rules approach, equations describing the physical behaviour of a system are used to predict the output. The output of the data-driven methods on the other hand is based on data of the historical behaviour of the system. The data-driven methods, which use regression models to find the most accurate function to map the input parameters to the observed output, can be further divided in statistical and machine learning methods. In statistics, the complexity of these functions is often predetermined by the regression model whereas in machine learning this complexity is learned by the method itself \cite{breiman}.


\section{Forward versus data-driven}
In the forward methods, the thermal load forecast of buildings is approximated based on the physical principles of the building. These equations can range from rather simple estimates of the thermal properties of the building materials up to detailed comprehensive building models \cite{alhomoud}. Developing detailed physical models of a building is often a costly and time-consuming activity because they require a lot of detailed system information and expert knowledge. Therefore many bulding design software tools, such as EnergyPlus \cite{energyplus} and TRNSYS \cite{trnsys}, already incorporate these complex energy simulation models. They are able to calculate heating and cooling loads as well as simulating energy consumption.

In contrast to forward methods, a data-driven approach creates a model describing the thermal load of a building based on available data of the historical behaviour of the building. Hence statistical and machine learning methods require collecting historical behavioural data. Statistical methods derive correlations between the target variable, e.g. the thermal load of the building, and influential parameters such as weather information and historic thermal load data. These methods are often used as baseline models for comparison with more elaborate methods. Several case studies indicate that they yield inferior prediction quality compared to machine learning approaches like artificial neural networks (ANNs) or support vector machines (SVMs) \cite{ekonomou,tso,azadeh1,azadeh2}. Neto et al. \cite{neto} compared the consumption forecasting capabilities of ANNs with the above mentioned EnergyPlus software based on a case study. The results showed that even though the EnergyPlus model was setup with very detailed information of the building under consideration, forecast errors were comparable with ANNs trained with 17 months of historic data. Together with the fact that forward methods are costly, time-consuming to develop and poorly generalizable we can conclude that data-driven methods are better suited for thermal load prediction in an operational context. The remainder of this paper will focus on several of these data-driven methods and more in particular on combining different types of data-driven methods in an expert system of thermal load forecasting experts to enhance the performance with respect to the best individual expert.

\section{\label{machine_learning}Machine learning}
Machine learning approaches are a subset of the above mentioned data-driven methods. They are used to devise complex models as well as prediction algorithms. In this paper the following techniques are applied: linear regression (LR), ANNs \cite{ann}, SVMs \cite{noble} and extremely randomized (extra) tree regressors (ETRs) \cite{wehenkel}. In the past years these methods have become increasingly popular techniques in forecasting energy consumption \cite{azadeh1,edwards,li,nasr,dong}. We will discuss these methods briefly in the next subsections, more information on them can be found in the papers referred to. Thereafter we will elaborate on creating an expert system combining these individual techniques.

\subsection{\label{lin_reg}Linear regression}
LR is an approach for modelling the relationship between a scalar dependent variable y and one or multiple explanatory variables denoted X. It is often used as a baseline model for the evaluation of machine learning methods, and we continue this practice analogous to the related studies \cite{ekonomou} and \cite{idowu}. We shall restrict attention to multiple linear regression (MLR) \cite{multiple_regression}, and model the thermal load $P$ at time $t$ by a linear equation.

\subsection{\label{anns}Artificial Neural Networks}
ANNs are inspired by the behaviour of biological neural networks, designed to simulate the way of how a human brain processes information \cite{ann}. They assemble their knowledge by detecting patterns and relationships in available data and learn through experience. Input values, e.g. historic thermal load information and weather data, are processed across the network topology by a number of weighted linear combinations and non-linear transformations to produce one or more output values, e.g. the forecasted thermal load. Azadeh et al. \cite{azadeh1} show the use of ANNs to predict electricity consumption in the Iranian agriculture sector, while in the work of Nasr et al. \cite{nasr} ANNs are used to forecast gasoline consumption in Lebanon. 

\subsection{\label{svm}Support vector machines}
A SVM is a computer algorithm that learns by example to assign labels to objects \cite{noble}. Besides classification, where the target variable takes class labels,  SVMs can also be applied for regression to enable continuous target values. The basic idea of an SVM is that a non-linear function is learned with an MLR, mapping the input variables to a higher dimensional feature space \cite{cortes}. According to Li et al. \cite{li}, who predicted the cooling load in Chinese office building, SVMs have a better prediction accuracy than a standard ANN. Dong et al. state a similar conclusion based on the prediction of energy consumption in several office buildings in Singapore. On the other hand Ekonomo \cite{ekonomou} carried out a performance comparison between the above mentioned LR, ANNs and SVMs based on predicting the energy demand in Greece. In this study the constructed ANN was more accurate than LR and had comparable performance to SVMs.

\subsection{\label{etr}Extremely randomized trees regressor}
In decision tree learning a decision tree is used as a predictive model which maps observations about an item (represented in the branches) to conclusions about the item's target value (represented in the leaves). Idowu et al. \cite{idowu} applied several machine learning methods on forecasting the thermal load in a DHS in Sweden and concluded that classical decision trees are outperformed by ANNs and SVMs, which makes them less suited for thermal load forecasting. In this paper however we use a tree-based ensemble method called extremely randomized trees or extra-trees \cite{wehenkel}. This algorithm averages predictions of a forest of trees obtained by partitioning the input-space with randomly generated splits, this leads to enhanced generalisation and reduced susceptibility to noise. Another advantage over classical trees and other ensemble methods is the reduced computational complexity of the extra-trees approach. In previous work the authors showed that extra-trees regressors are more accurate for forecasting thermal load than extreme-learning machines \cite{johansson}. Extreme learning machines are feed-forward neural networks with a single layer of hidden nodes, the weights between hidden nodes and outputs are learned in a single step while the weights between the input layer and the hidden nodes are chosen randomly \cite{ding}.

\section{\label{expert_advice}Expert advice}
Rather than comparing the performance of the individual methods described above the main goal of this study is to apply an algorithm able to combine $N$ thermal load forecasting experts, in a way that it always tracks the best of these $N$ experts. To achieve this we apply the concept of prediction with expert advice, this was first introduced by De Santis et al \cite{desantis}, in the meanwhile numerous studies on this topic have been published \cite{cesa1,vovk,littlestone,cesa2}. Prediction with expert advice allows to consider multiple stochastic models, each having different assumptions, in a single approach. We will discuss the concept of expert advice based on our case study of predicting the hourly thermal load in a DHS. 

Suppose that every day $k$ a forecaster wants to predict the hourly thermal load of the next 24 hours by combining a fixed set of individual experts. For this we thus have access to the predictions of a fixed and finite set of experts $\varepsilon = \left \{\varepsilon_1,\dots,\varepsilon_N \right \}$. On day $k$, all experts will make a thermal load estimate for the next 24 hours based on their available input data, e.g. historic thermal load and forecasted weather data. All experts will therefore return a vector $F_k = \left \{f_{1,k},\dots,f_{24,k} \right \}$ containing their hourly thermal load predictions for the next 24 hours. At each day $k$ the forecaster has access to the set $\left \{F_{E,k} : E \in \varepsilon \right \}$ representing the ``advice'' of each individual expert in the system . The forecaster then calculates the hourly thermal load $\hat{P}_k$ of the next 24 hours based on this set of information. At the end of the day, when the array of real thermal load values $Y_k$ is available, the experts' ``losses'' $\ell \left (F_{E,k}, Y_k \right )$ together with the forecaster's losses $\ell \left (\hat{P}_k, Y_k \right )$ are scored individually by a fixed loss function. This sequence is shown in algorithm  \ref{alg:ea}. 

\begin{algorithm}[t]
\caption{Prediction of thermal load with expert advice}\label{alg:ea}
\begin{algorithmic}[1]
\State Parameters: decision space  $\mathbb{R}_{\geq 0}$, outcome space $\mathbb{R}_{\geq 0}$, loss function $\ell$, set $\varepsilon$ of expert indices
\For{$k = 1,2,\dots$}
\State prediction of experts $\left \{F_{E,k} : E \in \varepsilon \right \}$, expert advice;
\State reveal expert advice to forecaster;
\State prediction of forecaster based on expert advice $\hat{P}_k$  
\State calculate forecaster's loss $\ell \left (\hat{P}_{k}, Y_k \right )$ and the expert losses $\ell \left (F_{E,k}, Y_k \right )$
\EndFor
\end{algorithmic}
\end{algorithm}

The difference between the forecaster's accumulated loss over day k ($\hat{L}_k$) and that of an expert i ($L_{i,k}$) is called regret. It measures how much the forecaster regrets, in hindsight, of not having followed the advice of a particular expert. Our goal is to find an algorithm having a small regret with regards to the best base expert in the class, this comes down to minimizing $R_k = \hat{L}_k - \displaystyle \min_{1 \leq i \leq N} L_{i,k}$. The next subsection elaborates on several solutions to achieve this goal.

\subsection{Aggregation rules}
In order to track the best expert using expert advice, Gaillard and Yannig \cite{gaillard} discussed four types of aggregation rules, applied on forecasting France's daily electricity consumption. Two of them are considered in our work, the fixed share forecaster (FS) and the polynomially weighted average forecaster with multiple learning rates (ML-Poly). Both are efficient generalized implementations of the exponentially weighted average forecaster (EWA) introduced by Littlestone and Warmuth \cite{littlestone} and by Vovk \cite{vovk}. The FS aggregation rules were introduced by Herbster and Warmuth \cite{herbster} while ML-poly was introduced by Gaillard et al. \cite{gaillard2} who add multiple learning rates to a version of the polynomially weighted average forecaster described by Cesa-Bianchi and Lugosi \cite{cesa3}. Both approaches add the notion of weights to algorithm \ref{alg:ea}. This implies that, based on the losses calculated in step 6, weights are assigned to the different base experts in order to combine the expert advice into the forecaster's prediction, calculated in step 5. These weights will minimize the forecaster's regret $R_k$. In the EWA concept, initially each expert has the same weight $w_i = 1/N, \forall i \in \left \{1,\dots,N \right \}$, after that the weight of an expert i at time t is calculated as follows:

\begin{equation}
w_{i,t} = \frac{e^{-\eta \sum\limits_{s=1}^{t-1} \ell \left (F_{i,s},Y_s \right )}}{\sum\limits_{n=1}^N e^{-\eta \sum\limits_{s=1}^{t-1} \ell \left (F_{n,s},Y_s \right )}}
\label{ewa_weight}
\end{equation}

Here $\eta$ represents a learning rate parameter which needs to be tuned. With proper tuning of $\eta$, EWA has a small average regret relative to the best fixed expert \cite{cesa1,devaine}. FS considers, besides $\eta$, also a mixing parameter $\alpha$ which takes into account the number of changes in the sequence of best experts. The initial weights should be chosen as $w_{1,0},\dots,w_{N,0} \geq 0$ such that $w_0 = w_{1,0}+\dots+w_{N,0} \leq 1$, this way a higher initial weight can be assigned to certain base experts in the class to increase their importance. After initialization, the weights will be calculated in two steps, first a new loss update is calculated:

\begin{equation}
v_{i,t} = \frac{w_{i,t-1}e^{-\eta \sum\limits_{s=1}^{t-1} \ell \left (F_{i,s},Y_s \right )}}{\sum\limits_{n=1}^N w_{i,t-1} e^{-\eta \sum\limits_{s=1}^{t-1} \ell \left (F_{n,s},Y_s \right )}}
\label{fs_loss_update}
\end{equation}

Here the weights of round $t-1$ are used together with the accumulated loss of each expert up to round $t-1$ and learning rate $\eta$ to calculate the updates. Once these updated losses are calculated, $\alpha$ is introduced to calculate the new weights of the experts:

\begin{equation}
w_{i,t} = \frac{\alpha}{N}+\left (1-\alpha \right ) v_{i,t} \textnormal{ where } \alpha \in [0,1]
\label{fs_weight}
\end{equation}

A positive value of $\alpha$ ensures that every expert has a minimal weight, which enables tracking the best compound action. Choosing $\alpha = 0$ will reduce the weights of the FS approach to $w_{i,t} = v_{i_t}$ which is equal to the EWA forecaster. For performance comparison we also implemented a version of the ML-Poly forecaster in which the multiple learning parameters are theoretically fixed and do not need to be tuned to the application. In our study the results of the ML-Poly forecaster are as good as the FS forecaster, which confirms the analysis made in \cite{gaillard}. For the sake of simplicity we will discuss the performance of our expert system based on the results obtained by the FS forecaster. However, more information on the ML-Poly forecaster can be found in \cite{gaillard} and \cite{gaillard2}.

\section{\label{case_study}Case study}
For this paper, the DHS in Rottne, Sweden, was used as a case study. This is a traditional 3\textsuperscript{rd} generation DHS located in the south of Sweden and operated by V\"axj\"o Energi. The piping in the network is about 10 300 m in length, with a total volume of about 64 m\textsuperscript{3}. The production units became operational in 1998 and at the time it consisted of a 1.5 MW burner for dry wood fuel and a 3 MW fossil oil burner. In 2004, the wood burner was refitted to work with more moist wood chip fuels, which lowered the capacity to 1.2 MW. Another biomass burner for wood chip fuels was also installed with a capacity of 1.5 MW. In 2012, the oil burner was upgraded to facilitate the use of biodiesel instead of fossil oil. Since then all heat generation in the Rottne district heating system is based on renewables. 

There are about 200 buildings connected to the district heating system in Rottne, and of these about 150 are single-family domestic dwellings. The rest are connections to commercial customers, and of those the ten largest consumers are connected to the system used in this study. These ten controllable customers represent nearly one-third of the total heat demand, including distribution losses, in the DHS. Each such building is fitted with a district heating substation controlled by an existing controller. During the project a retrofit device was added to this existing controller hardware, which makes it possible to interact with it remotely through an outdoor temperature sensor override mechanism. This makes it possible to control the substation by sending alternative outdoor temperature signals, which the underlying controller will then respond to according to its default settings. Additional sensors were also added to measure temperature data from the supply and return temperature on the heating system side, as well as to read data from the heat meter. 

Although biodiesel is considered renewable it is still quite expensive, and its use should be avoided if at all possible. The operational behaviour of the production units is connected to the supply temperature levels at the production site. If the combined production units are not able to generated the required heat demand, the supply temperature will start to drop. This, in turn, will trigger the diesel burner. This normally happens at a thermal load of about 2.7 MW. Therefore, it is of great importance for the operational optimisation mechanisms to be able to forecast when the system is about to reach such levels, since this facilitates the use of demand side management to minimise or avoid such peak loads. 

\section{\label{input_data}Available input data}
The input data used in this study is collected directly from the IT platform implemented at the Rottne DHS. This platform enables us to retrieve operational data of the DHS in real-time together with historic control signals, thermal load and weather forecast data. Thermal load data and control signals are available on quarter-hourly basis while weather forecasts consist of hourly values. The heat that the buildings extract from the DHS is controlled based on a heating curve. This is a common rule-based control concept which will increase the load of a building when the outdoor temperature decreases. As a result, one can manipulate the heat consumption of a building for a short period of time, typically minutes up to a couple of hours, by adapting the outdoor temperature measurement. As stated in section \ref{case_study}, a network controller is installed which will manipulate these outdoor temperature measurements to enable automatic demand side management. More concretely, the control signal of this network controller consists of the difference between the real outdoor temperature and a fake virtual one. We will therefore refer to these control signals as delta T values. The total raw dataset spans a period of 27 months, from November 2014 up to February 2017. 

\subsection{\label{dataset_analysis}Dataset analysis and feature selection}
It is of great importance to analyse the relationships between the variables available in the dataset and the target value, e.g. the thermal load. These relationships will enable us to select the appropriate features to use in the forecasting techniques discussed in section \ref{machine_learning}.

\begin{figure}
  \centering
  \begin{tabular}{@{}c@{}}
    \includegraphics[scale=0.35]{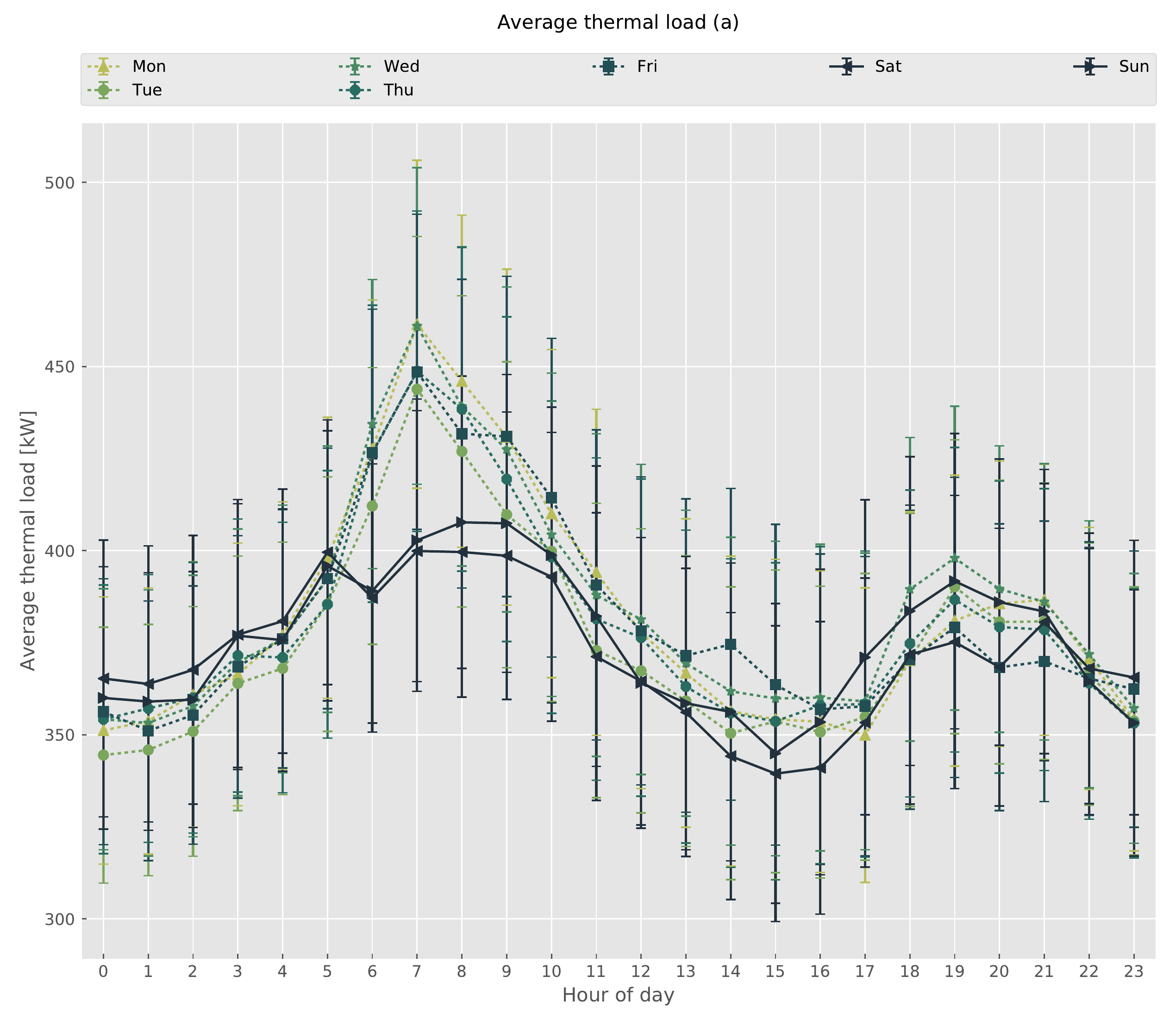}
  \end{tabular}

  \begin{tabular}{@{}c@{}}
    \includegraphics[scale=0.35]{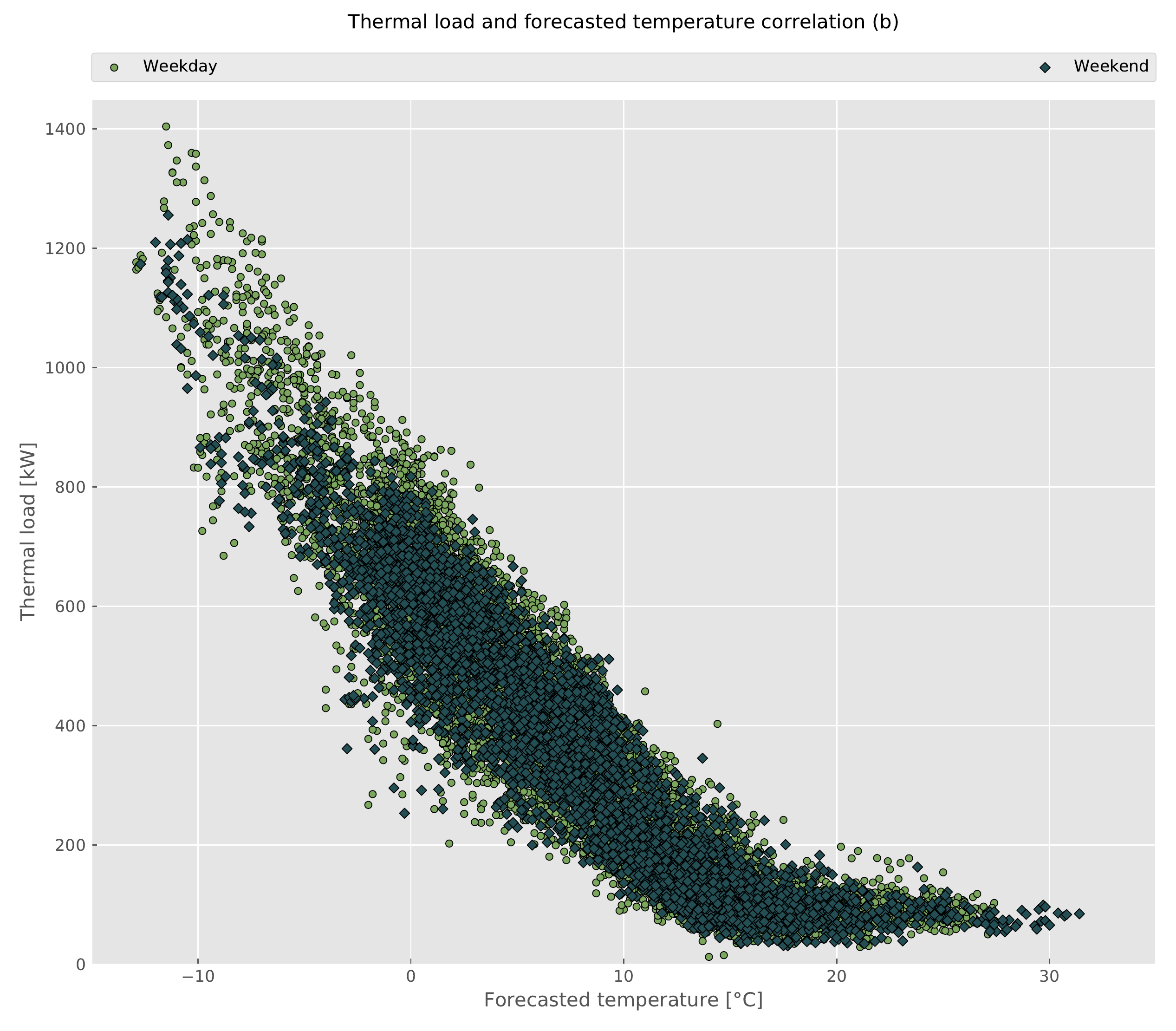}
  \end{tabular}

\caption{\label{avg_heat_load_and_corr} (a) The mean and confidence interval of the thermal load per day of the week and hour of the day. (b) The correlation between the thermal load and the forecasted outdoor temperature during the week and in the weekend.}
\end{figure}

Figure \ref{avg_heat_load_and_corr}a shows an analysis of the hourly thermal load based on the mean and 90\% confidence interval with regards to the day of the week. It is apparent that the morning peak load during weekdays is significantly higher than the morning peak in the weekends, furthermore we also see the logical thermal load dependency on the time of the day. The Pearson product moment correlation coefficient (PPMCC), shown in graph \ref{avg_heat_load_and_corr}a, was used to determine the relationship between the forecasted outdoor temperature and the thermal load. The PPMCC of -0.92 shows a strong negative linear correlation, especially for outside temperature forecasts below 15 \degree C. From graph \ref{avg_heat_load_and_corr}b it is also confirmed that in low temperature operation, the peak load is higher on weekdays than in weekends. In order to capture the above relationships, the following features will be included: day of the week, hour of the day and forecasted outdoor temperature.

\begin{figure}
	\centering
	\includegraphics[scale=0.38]{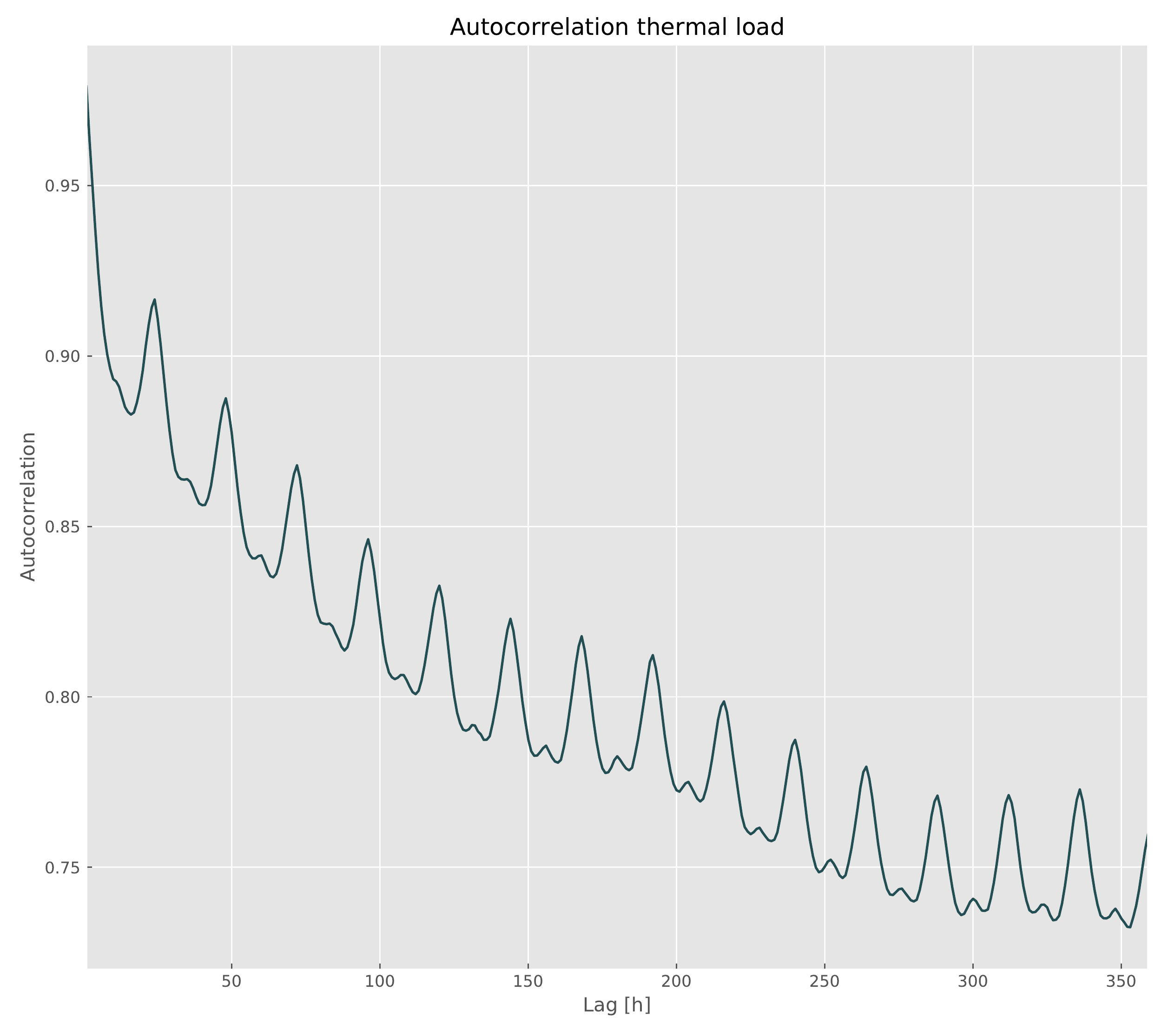} 
	\caption{\label{autocorr} Autocorrelation of the thermal load}
\end{figure}

The result of the thermal load autocorrelation is shown in figure \ref{autocorr}. It can be seen that the thermal load is also highly correlated with the thermal load lagging a multiple of 24 hours. The correlation drops the further we go back in time. To take into account the historic thermal load and the day of the week variations, the following features are added to the set: the thermal load with a lag of 1 day (24 hours) and 1 week (168 hours). Because the thermal load is also influenced by the control signals (dT) these too are included in the feature set with the same lag, 24 hours and 168 hours, as the thermal load.

Table \ref{feature_sets} gives an overview of the different feature sets we applied in our expert system. The first one is the full feature set as described above which takes into account timing information, temperature forecast, historic thermal loads and control signals. The second set does not contain the historic control signals (dT) and the third set does not contain historic control signals nor historic thermal load. All these data sets are split into a training and test set. We first use the training set to optimize the hyper-parameters of the models by means of cross-validation \cite{cross_validation}, more information on this can be found in section \ref{implementation}. Thereafter, the same training set is used to train the models by pairing the input with the known expected target value. In the end, the models included in our expert system are applied to the test set to asses their performance. The 27 months long dataset is split up according to the 75/25 principle, this results in 20 months of training data (November 2014 to July 2016), and 7 months of test data (August 2016 to February 2017). In the next section we will discuss which feature sets are used in combination with the different experts to analyse the impact on the expert system's performance.

\begin{table}
\centering
\caption{Feature sets with Hour of Day (HoD), Day of Week (DoW), Day of Year (DoY), forecasted outdoor temperature ($\hat{T}_{out}$), yesterday's thermal load (P\textsubscript{t-24}), last week's thermal load (P\textsubscript{t-168}), yesterday's control signal (dT\textsubscript{t-24}) and last week's control signal (P\textsubscript{t-168}) }
\label{feature_sets}
\begin{tabular}{l|ccc|c|cc|cc}
& \multicolumn{3}{c|}{Timing}         & Temp  & \multicolumn{2}{c|}{Thermal load} & \multicolumn{2}{c}{Control signal} \\
              & HoD & DoW & DoY & $\hat{T}\textsubscript{out}$ & P\textsubscript{t-24}              & P\textsubscript{t-168}             & dT\textsubscript{t-24}              & dT\textsubscript{t-168}              \\ \hline
Full set      & \cmark           & \cmark           & \cmark           & \cmark             & \cmark                 & \cmark & \cmark                  & \cmark              \\ \hline
Set-dT   & \cmark           & \cmark           & \cmark           & \cmark             & \cmark                 & \cmark                 & \xmark                  & \xmark                    \\ \hline
Set-lags & \cmark           & \cmark           & \cmark           & \cmark             & \xmark                 & \xmark                 & \xmark                  & \xmark                  
\end{tabular}
\end{table}

\section{\label{implementation_results}Implementation and results}
First we will elaborate on the detailed implementation of the forecasting methods discussed in section \ref{machine_learning}, secondly we will discuss the results obtained from applying these methods, in combination with the expert system, to the dataset discussed in section \ref{input_data}. It is not our goal to go into all the details of the underlying machine learning methods. However, to provide information up to a level that is necessary for reproducing the results, the use of some machine learning jargon is inevitable. More information on this technical jargon can be found in the numerous references given throughout the discussion.

\subsection{\label{implementation}Implementation}
All the code is implemented in Python 2.7 and 3.5 \cite{python} using the machine learning library scikit-learn \cite{scikit-learn}, version 0.17.1 and 0.18.1. For every forecaster the scikit-learn API offers a simple fit method to train the forecaster and a predict method to forecast the target values. The training set will serve as an input to the fit method while results will be obtained by providing the test set to the predict method.

The first and most straightforward method, LR (section \ref{lin_reg}), is implemented by using the built-in scikit-learn's LinearRegression. No extra parameters have to be defined before training the regressor. We integrated one LR expert, serving as a baseline scenario, which is trained using the full feature set.

Secondly the ANNs (section \ref{anns}) are implemented using the KerasRegressor functionality. Keras is a neural networks library for Python \cite{keras} capable of using the TensorFlow \cite{tensorflow} and Theano \cite{theano} backend. Both backends are open-source software libraries able to build neural networks. In this implementation the TensorFlow backend is used but a comparison showed identical results when using Theano. Based on the outcome of several experiments on our dataset we constructed an ANN with two hidden layers each having twelve hidden units, no dropout or regularisation \cite{srivastava} is used because the ANN is rather small. A rectified linear unit (ReLU) \cite{relu}, the most popular activation function in deep neural networks, is used in both hidden layers. We use cross-validation to find the optimal hyper-parameters, number of epochs and batch size, of the ANN \cite{cross_validation}. Cross-validation is a technique to evaluate predictive models by partitioning the original set repeatedly into a training set to train the model, and a test set to evaluate it. Here we execute an exhaustive search over the specified hyper-parameter values by applying scikit-learn's GridSearchCV on the training set. GridSearchCV will repeatedly split the full training set into a subset for training and a subset for testing to evaluate the hyper-parameters. In our case this resulted in initializing the ANN with 200 epochs and a batch size of 10. Once the above parameters have been tuned, the ANN is trained with the training set. Two ANNs are included in our expert system, one which is trained with the full feature set and another one which is trained without the historic control signals. 

Thirdly the SVMs are implemented using the Epsilon-support vector regression ($\epsilon$-SVR) \cite{svr}. The goal of this $\epsilon$-SVR is to find a function $f(x)$ that deviates at most $\epsilon$ from the observed target value and at the same time is as simple as possible. A radial basis function (RBF) \cite{rbf} is chosen as the kernel for the $\epsilon$-SVR. As with the ANNs, GridSearchCV is applied to find the optimal values for the regularization parameter $C$, the kernel parameter $\gamma$ and $\epsilon$. This $C$ parameter represents a trade-off between misclassification of training samples and the complexity of the decision surface. A high $C$ value can lead to overfitting while a low $C$ value can lead to underfitting. The $\gamma$ parameter on the other hand defines how far the influence of one training sample reaches. High values of $\gamma$ result in a close reach, possibly ending up in overfitting and low values of $\gamma$ result in a far reach, possibly ending up in underfitting. The $\epsilon$ parameter specifies the margin in which no penalty is given to points predicted within a distance $\epsilon$ from the actual training value. Based on the GridSearchCV results, the SVMs used in our expert system are initialized with $C = 1000$, $\gamma = 0.00001$ and $\epsilon = 0.01$. We added two SVMs to our expert system, trained with training sets analogue to the ANNs discussed above.

Lastly the ETRs (section \ref{etr}) are implemented using the ExtraTreeRegressor functionality. Again GridSearchCV was used to determine the optimal hyper-parameters. Three ETRs are added to our expert system, each having 100 trees in their forest and a minimum of 7 samples per leaf. The first one is trained with the full set of features, the second one is trained without historic control signals and the third one is trained without historic control signals nor historic thermal load information. The minimum number of samples to split an internal node of a tree is equal to the number of features the tree is trained with plus one.

Table \ref{experts} gives an overview of the eight different estimators contained in the expert system.

\begin{table}
\centering
\caption{Overview of the experts included in the expert system with linear regression (LR), artificial neural network (ANN), support vector machine (SVM) and extra-trees regressor (ETR)}
\label{experts}
\begin{tabular}{l|c|c|c}
                 & Full set & Set-dT & Set-lags \\ \hline
LR				 & \cmark                       & \xmark           & \xmark   \\ \hline
ANN              & \cmark                       & \cmark           & \xmark   \\ \hline
SVM              & \cmark                       & \cmark           & \xmark   \\ \hline
ETR              & \cmark                       & \cmark           & \cmark            
\end{tabular}
\end{table}

\subsection{\label{results}Results and discussion} 
The performance assessment of the individual experts as well as the global forecaster is carried out by comparing the mean absolute percentage error (MAPE), as defined in equation \ref{mape}, based on an hourly forecast for a 24 hour horizon.

\begin{equation}
MAPE = \frac{100}{n} \sum\limits_{t=1}^{n} \left| \frac{A_{t} - F_{t}}{A_{t}} \right|
\label{mape}
\end{equation}

With $F_{t}$ the predicted value, $A_{t}$ the real value and n the number of predictions. 

\begin{figure}
	\centering
	\includegraphics[scale=0.38]{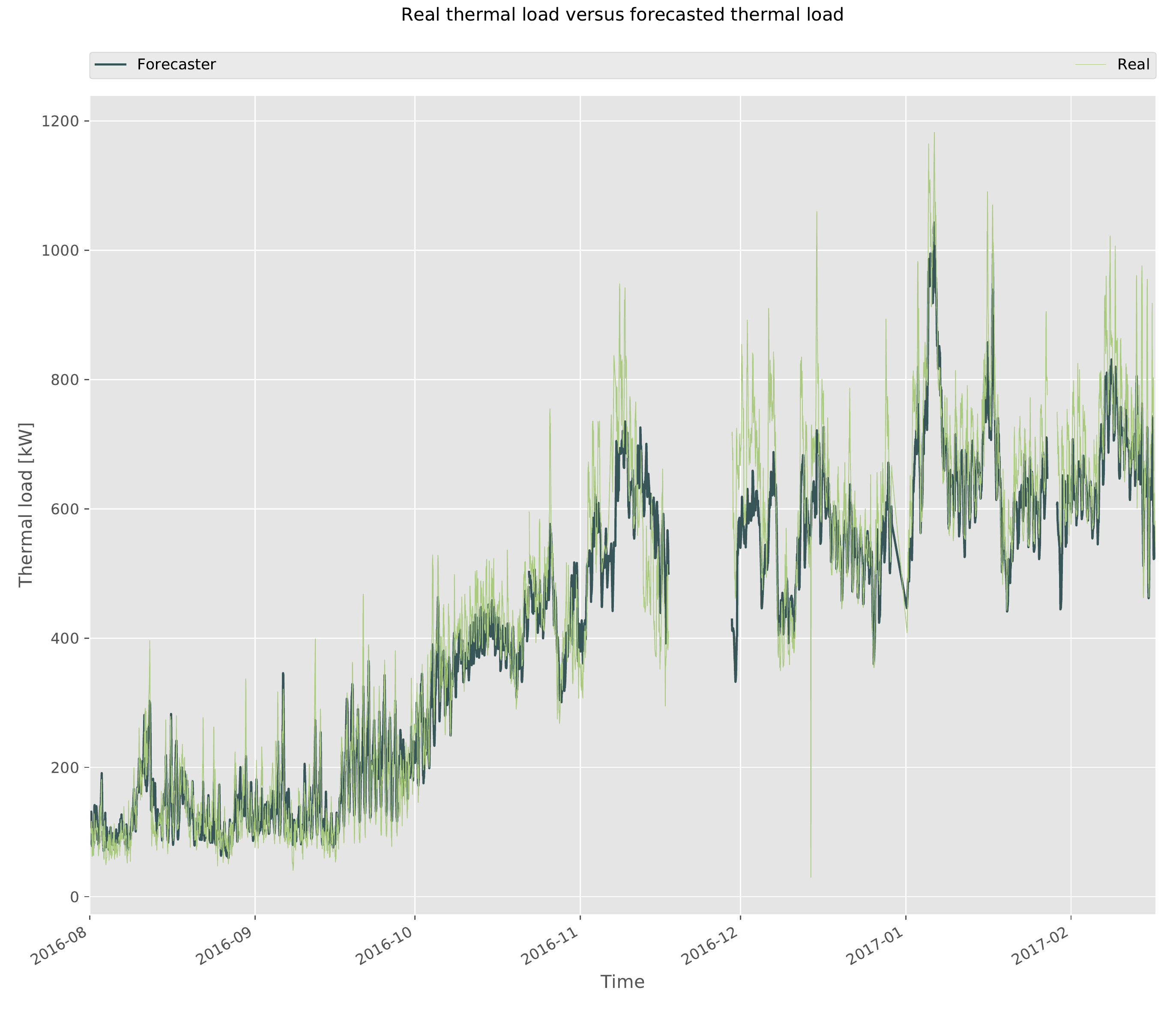} 
	\caption{\label{forecast} Real thermal load versus forecasted thermal load}
\end{figure}

All eight experts are trained with training data from November 2014 to July 2016. Subsequently, every expert calculates a day by day forecast of the thermal load of the next day, hereby spanning the complete time range of the test set (August 2016 to February 2017). Each day the experts are scored based on the method described in section \ref{expert_advice}, using MAPE as the loss function. Thereafter, the weights of the experts are updated accordingly. As there are eight experts in our system, each one starts with an initial weight of $w = 1/8$. The forecaster then  predicts the hourly thermal load of the next day by combining the weighted predictions of the individual experts. In figure \ref{forecast} this forecasted thermal load is plotted together with the real heat load of the DHS over the complete time range of the test set. No data is available from 17 November 2016 up to 29 November 2016 due to a technical failure in the IT infrastructure of the system.

\begin{figure}
  \centering
  \begin{tabular}{@{}c@{}}
    \includegraphics[scale=0.35]{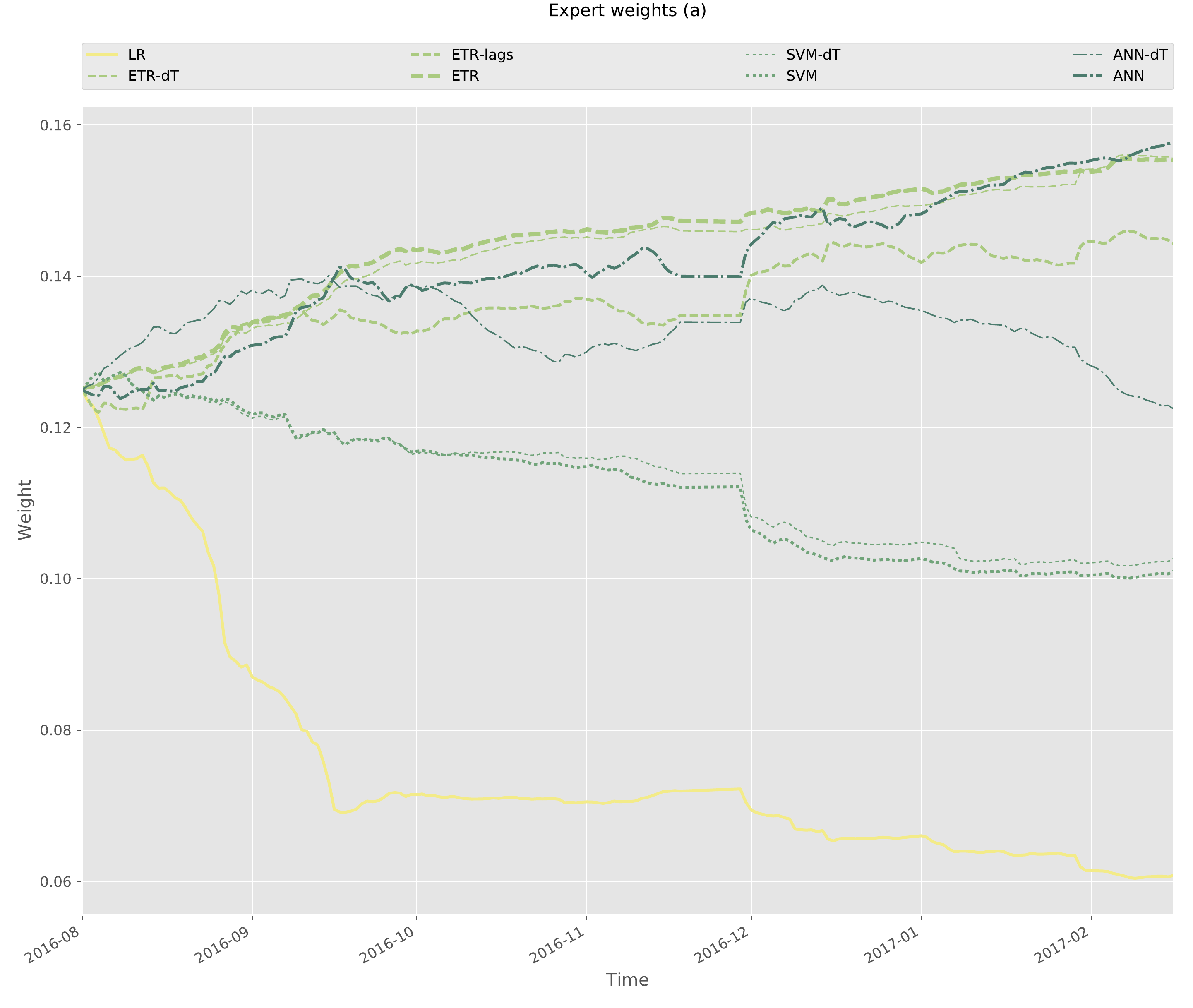}
  \end{tabular}

  \begin{tabular}{@{}c@{}}
    \includegraphics[scale=0.35]{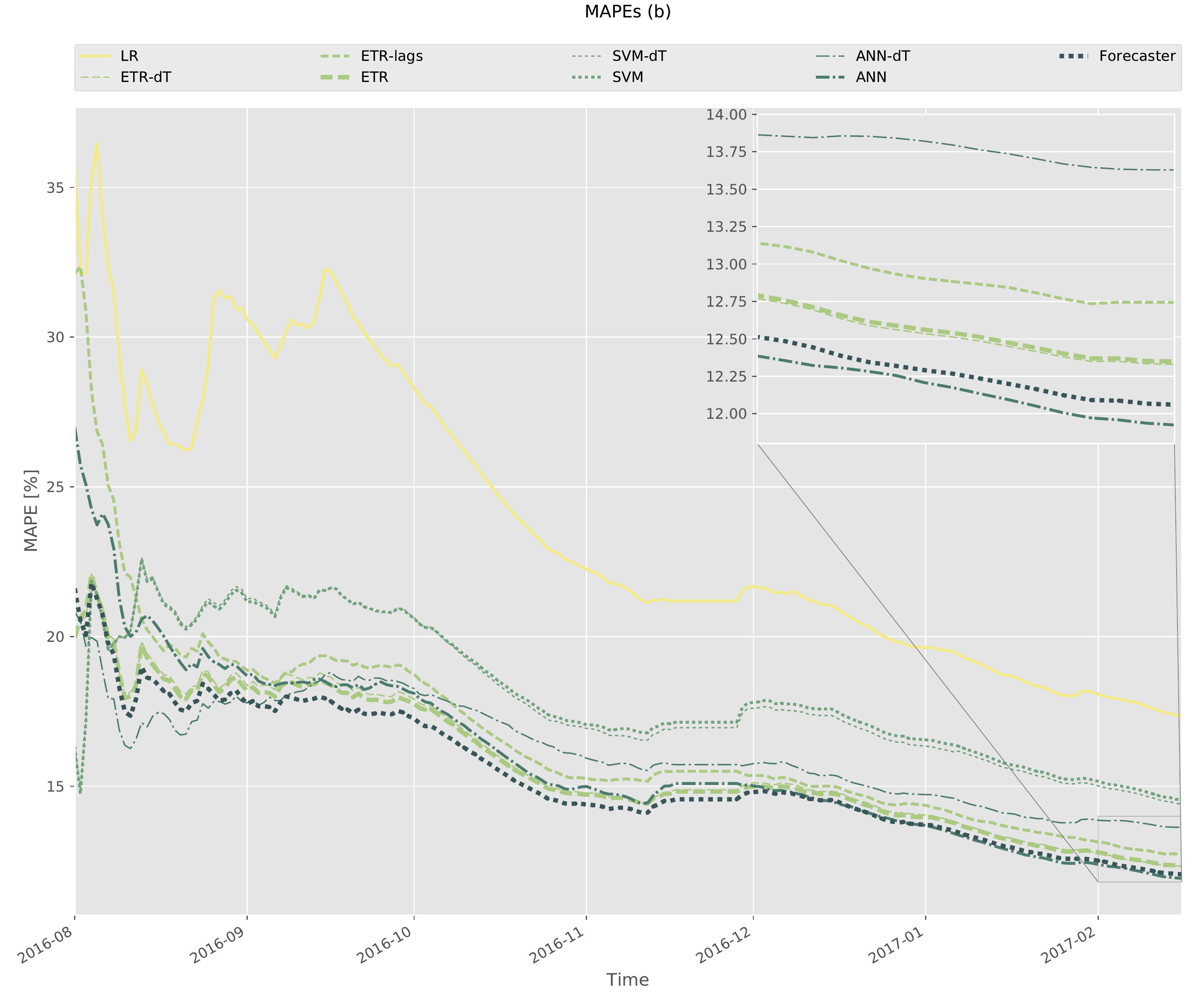}
  \end{tabular}

\caption{\label{weights_mapes} Expert weights (a) and moving average of MAPEs (b)}
\end{figure}

Figure \ref{weights_mapes}a shows the evolution of the individual expert's weights over time while figure \ref{weights_mapes}b shows the moving average of the expert's MAPEs. It is clear that the LR performs substantially worse than all the other experts, therefore it always has the lowest weight. It also stands out that the ANN with the full feature set has the best overall performance, followed by the three ETRs. Furthermore, it is interesting to see the performance increase with increased thermal loads, as seen from October 2016 up to February 2017. As the expert system will be used as part of a control system responsible for limiting high thermal loads in the DHS, this is a helpful property of the forecaster. The compound prediction, presented by the forecaster label in graph \ref{weights_mapes}b, performs almost exactly the same as the best forecaster in the system having a MAPE of 12.06\%. However if we only take into account the months with high thermal loads, October 2016 to February 2017, the forecaster even obtains a MAPE of 9.75\%. 

To obtain the results discussed above the individual forecasters were only trained once, using the available training set of November 2014 to July 2016. The test set however ranges from August 2016 to February 2017. In order to capture potential changes in the DHS or the controllable buildings, we retrained the experts daily by adding the previous day of the test set to the training set. Table \ref{retrain} presents a comparison of the MAPEs without retraining (top) and the MAPEs with daily retraining (bottom). The table only shows the results of the experts trained with the complete feature set. It is apparent that the results are almost identical, from this we can conclude that no changes took place in the DHS or buildings during the test set.

\begin{table}
\centering
\caption{\label{retrain} Comparison of of the MAPEs  with no retraining (top) and daily retraining (bottom) of the experts}
\begin{tabular}{c|c|c|c|c|c}
              & LR    & ETR   & SVM   & ANN  & Forecaster \\ \hline
No retrain    & 17.34 \% & 12.34 \% & 14.54 \% & \textbf{11.92 \%} & 12.06 \%     \\ \hline
Daily retrain & 17.27 \%  & 12.42 \%  & 14.72 \%  & \textbf{11.56 \%}  & 11.95 \%
\end{tabular}
\end{table}

\section{\label{conclusion}Conclusion}
A fixed share forecaster expert system, combining the outcome of eight individual experts into one forecast, is presented in this paper. The experts differentiate in both machine learning approach and number of features used to train them. The following techniques are used: LR, ANNs, SVMs and ETRs in combination with three different feature sets, a first one solely taking into account timing information and an outdoor temperature forecast, the second one adding historic thermal load information and the last one adding historic control signals. To train the system and analyse its performance a dataset of 27 months was used ranging from November 2014 up to February 2017. The first 20 months were used as a training and validation set while the last 7 months were used as a test set performance analysis. Of all the experts in the system, the LR performs worst while the ANNs and ETRs are slightly better than the SVMs. From the retraining interval analysis we concluded that, in this case study, retraining does not increase the forecaster's performance. This is due to the extensive initial training set together with a DHS that did not change throughout the test set. Over this test set, the expert system achieves our predefined goal of tracking the best expert, the ANN with full feature set, in the system. Beyond this, combining different experts adds robustness to the forecaster and reduces susceptibility to changes in the DHS. Our implementation also allows for easy integration of new experts as long as they provide the fit and predict interface given by scikit-learn. Future research will consist of integrating this expert system in a DHS control solution using the thermal load forecast for peak shaving. It will enable us to implement an automatic demand response system able to control a number of buildings in the DHS to limit the thermal peak load in order to avoid the use of biodiesel burners.

\section{\label{Acknowledgement}Acknowledgement}
This work has been carried out by Noda, EnergyVille/VITO and V\"axj\"o Energi within the context of the STORM project, funded by the European Union's Horizon 2020 programme under grant agreement 649743.

\section*{References}

\bibliography{mybibfile}

\end{document}